\documentclass[runningheads]{llncs}
\usepackage[T1]{fontenc}
\usepackage{graphicx}
\usepackage{makecell}
\usepackage{multirow}
\usepackage{amsmath}
\usepackage{amsfonts}
\usepackage{mathrsfs}
\usepackage[bookmarks=true]{hyperref}
\usepackage{lipsum}
\usepackage{xcolor}
\usepackage{bbding}

\definecolor{gray}{RGB}{128,128,128}
\SetLipsumParListSurrounders{\begingroup\color{gray}}{\endgroup}

\definecolor{myred}{RGB}{255,0,0}
\definecolor{mygreen}{RGB}{0,176,80}

\usepackage{xspace}
\makeatletter
\DeclareRobustCommand\onedot{\futurelet\@let@token\@onedot}
\def\@onedot{\ifx\@let@token.\else.\null\fi\xspace}

\def\ie{\emph{i.e}\onedot}

\makeatother

\usepackage{color}

\begin{document}
\title{Devil is in Channels: Contrastive Single Domain Generalization for Medical Image Segmentation}
\titlerunning{Contrastive Single Domain Generalization for Medical Image Segmentation}
\author{Shishuai Hu\inst{1} \and
Zehui Liao\inst{1} \and
Yong Xia\inst{1,2,3}\Envelope}
\authorrunning{S. Hu et al.}
\institute{
National Engineering Laboratory for Integrated Aero-Space-Ground-Ocean Big Data Application Technology, School of Computer Science and Engineering, Northwestern Polytechnical University, Xi’an 710072, China \\
\email{yxia@nwpu.edu.cn}
\and
Ningbo Institute of Northwestern Polytechnical University, Ningbo 315048, China
\and
Research and Development Institute of Northwestern Polytechnical University in Shenzhen, Shenzhen 518057, China
}
\maketitle              % typeset the header of the contribution
%
% !TEX root = paper.tex

\begin{abstract}
Deep learning-based medical image segmentation models suffer from performance degradation when deployed to a new healthcare center.
To address this issue, unsupervised domain adaptation
and multi-source domain generalization
methods have been proposed, which% Both settings
, however, are less favorable for clinical practice due to the cost of acquiring target-domain data and the privacy concerns associated with redistributing the data from multiple source domains.
In this paper, we propose a \textbf{C}hannel-level \textbf{C}ontrastive \textbf{S}ingle \textbf{D}omain \textbf{G}eneralization (\textbf{C$^2$SDG}) model for medical image segmentation.
In C$^2$SDG, the shallower features of each image and its style-augmented counterpart are extracted and 
used for contrastive training, resulting in the disentangled style representations and structure representations. The segmentation is performed based solely on the structure representations.
Our method is novel in the contrastive perspective that enables channel-wise feature disentanglement using a single source domain.
We evaluated C$^2$SDG against six SDG methods on a multi-domain joint optic cup and optic disc segmentation benchmark. 
Our results suggest the effectiveness of each module in C$^2$SDG and also indicate that C$^2$SDG outperforms the baseline and all competing methods with a large margin.
The code will be available at \url{https://github.com/ShishuaiHu/CCSDG}.

\end{abstract}
\section{Introduction}
It has been widely recognized that the success of supervised learning approaches, such as deep learning, relies on the i.i.d. assumption for both training and test samples~\cite{LITJENS201760}.
This assumption, however, is less likely to be held on medical image segmentation tasks due to the imaging distribution discrepancy caused by non-uniform characteristics of the imaging equipment, inconsistent skills of the operators, and even compromise with factors such as patient radiation exposure and imaging time~\cite{IMGCHAR}.
Therefore, the imaging distribution discrepancy across different healthcare centers renders a major hurdle that prevents deep learning-based medical image segmentation models from clinical deployment~\cite{guan2021domain,xie_survey_2021}.

To address this issue, unsupervised domain adaptation (UDA)~\cite{wilson2020survey,hu_domain_2022} and multi-source domain generalization (MSDG)~\cite{hu2022domain,wang2022generalizing} have been studied. UDA needs access to the data from source domain(s) and unlabeled target domain, while MSDG needs access to the data from multiple source domains. In clinical practice, both settings are difficult to achieve, considering the cost of acquiring target-domain data and the privacy concerns associated with redistributing the data from multiple source domains~\cite{zhou2022domain,hu2022prosfda}. 

\begin{figure}[t]
  \centering
  \includegraphics[width=1\textwidth]{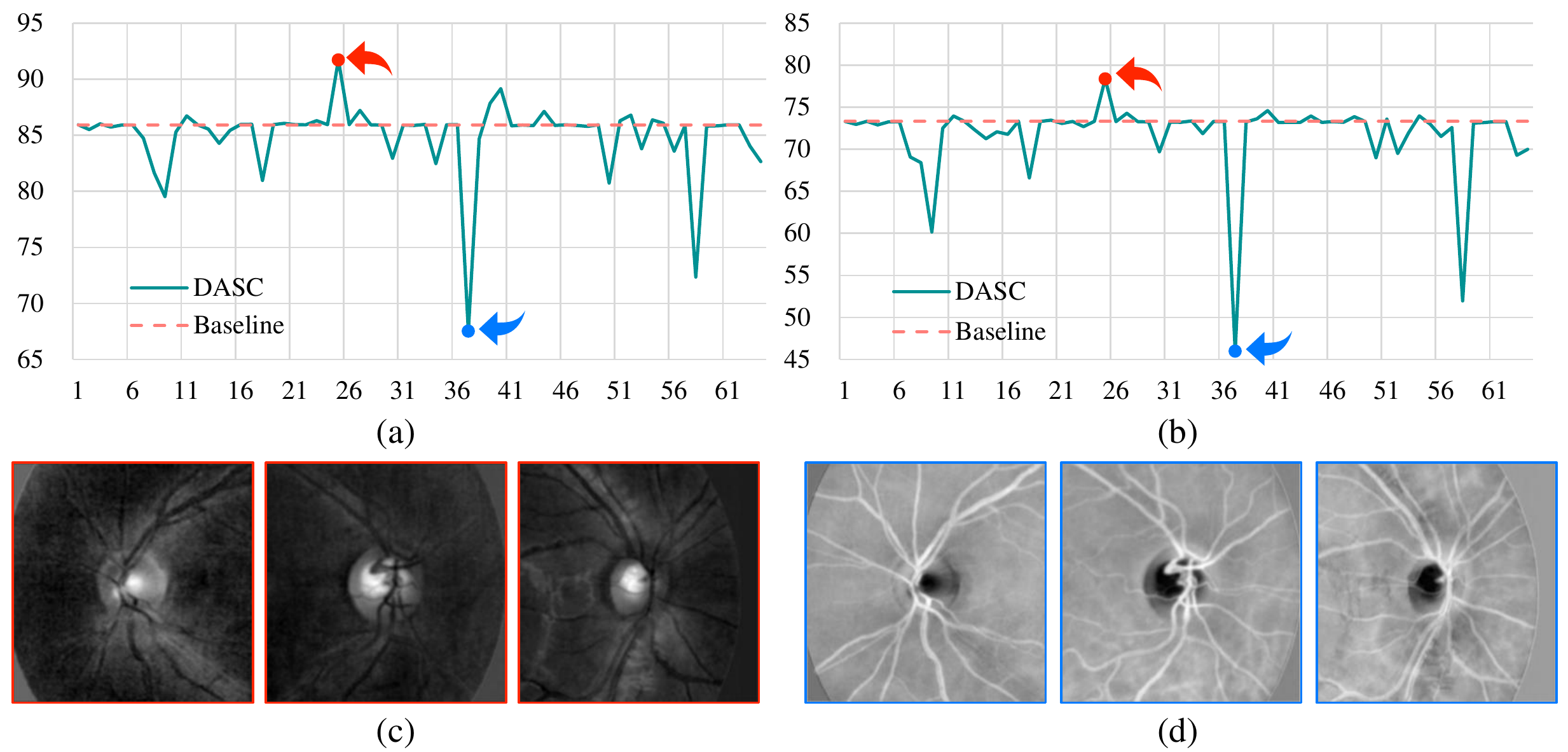}
  \caption{Average OD (a) and OC (b) segmentation performance (Dice\%) obtained on unseen target domain (BASE2) versus removed channel of shallow features. The Dice scores obtained before and after dropping a channel are denoted by `Baseline' and `DASC', respectively. The 24th channel (c) and 36th channel (d) obtained on three target-domain images are visualized.}
  \label{fig:motivation}
\end{figure}

By contrast, single domain generalization (SDG)~\cite{zhou2022domain,su2022rethinking,xu2022adversarial,chen2022maxstyle,zhou2022generalizable,ouyang2022causality} is a more practical setting,
under which only the labeled data from one source domain are used to train the segmentation model, which is thereafter applied to the unseen target-domain data.
The difficulty of SDG is that, due to the existence of imaging distribution discrepancy, the trained segmentation model is prone to overfit the source-domain data but generalizes poorly on target-domain data.
An intuitive solution is to increase the diversity of training data by performing data augmentation at the image-level~\cite{zhang2020generalizing,su2022rethinking,xu2022adversarial,ouyang2022causality}.
This solution has recently been demonstrated to be less effective than a more comprehensive one, \ie, conducting domain adaptation on both image- and feature-levels~\cite{ma2022i2f,chen2022maxstyle,hu_domain_2022}.
As a more comprehensive solution, Dual-Norm~\cite{zhou2022generalizable} first augments source-domain images into `source-similar' images with similar intensities and `source-dissimilar' images with inverse intensities, and then processes these two sets of images using different batch normalization layers in the segmentation model.
Although achieving promising performance in cross-modality CT and MR image segmentation, Dual-Norm may not perform well under the cross-center SDG setting, where the source- and target-domain data are acquired at different healthcare centers, instead of using different imaging modalities. In this case,  the `source-dissimilar' images with inverse intensities do not really exist, and it remains challenging to determine the way to generate both `source-similar' and `source-dissimilar' images~\cite{decenciere2014feedback,almazroa2018retinal}.
To address this challenge, we suggest resolving `similar' and `dissimilar' from the perspective of contrastive learning.
Given a source image and its style-augmented counterpart, only the structure representations between them are `similar', whereas their style representations should be `dissimilar'.
Based on contrastive learning, we can disentangle and then discard the style representations, which are structure-irrelevant, using images from only a single domain.

Specifically, to disentangle the style representations, we train a segmentation model, \ie, the baseline, using single domain data and assess the impact of the features extracted by the first convolutional layer on the segmentation performance, since shallower features are believed to hold more style-sensitive information~\cite{xie_survey_2021,hu_domain_2022}.
A typical example was given in~\figurename{~\ref{fig:motivation}} (a) and (b), where the green line is the average Dice score obtained on the target domain (the BASE2 dataset) versus the index of the feature channel that has been dropped. 
It reveals that, in most cases, removing a feature does not affect the model performance, indicating that the removed feature is redundant. For instance, the performance even increases slightly after removing the 24th channel. This observation is consistent with the conclusion that there exists a sub-network that can achieve comparable performance~\cite{frankle2018lottery}.
On the contrary, it also shows that some features, such as the 36th channel, are extremely critical. Removing this feature results in a significant performance drop.
We visualize the 24th and 36th channels obtained on three target-domain images in~\figurename{~\ref{fig:motivation}} (c) and (d), respectively.
It shows that the 36th channel is relatively `clean' and most structures are visible on it, whereas the 24th channel contains a lot of `shadows'.
The poor quality of the 24th channel can be attributed to the fact that the styles of source- and target-domain images are different and the style representation ability learned on source-domain images cannot generalize well on target-domain images. 
Therefore, we suggest that the 24th channel is more style-sensitive, whereas the 36th channel contains more structure information.
This phenomenon demonstrates that `the devil is in channels'. Fortunately, contrastive learning provides us a promising way to identify and expel those style-sensitive `devil' channels from the extracted image features.

In this paper, we incorporate contrastive feature disentanglement into a segmentation backbone  
and thus propose a novel SDG method called \textbf{C}hannel-level \textbf{C}ontrastive \textbf{S}ingle \textbf{D}omain \textbf{G}eneralization (\textbf{C$^2$SDG}) for joint optic cup (OC) and optic disc (OD) segmentation on fundus images.
In C$^2$SDG, the shallower features of each image and its style-augmented counterpart are extracted and used for contrastive training, resulting in the disentangled style representations and structure representations. The segmentation is performed based solely on the structure representations.
This method has been evaluated against other SDG methods on a public dataset and improved performance has been achieved.
Our main contributions are three-fold:
(1) we propose a novel contrastive perspective for SDG, enabling contrastive feature disentanglement using the data from only a single domain;
(2) we disentangle the style representations and structure representations explicitly and channel-wisely, and then diminish the impact of style-sensitive `devil' channels; and 
(3) our C$^2$SDG outperforms the baseline and six state-of-the-art SDG methods on the joint OC/OD segmentation benchmark.

\begin{figure}[t]
  \centering
  \includegraphics[width=1\textwidth]{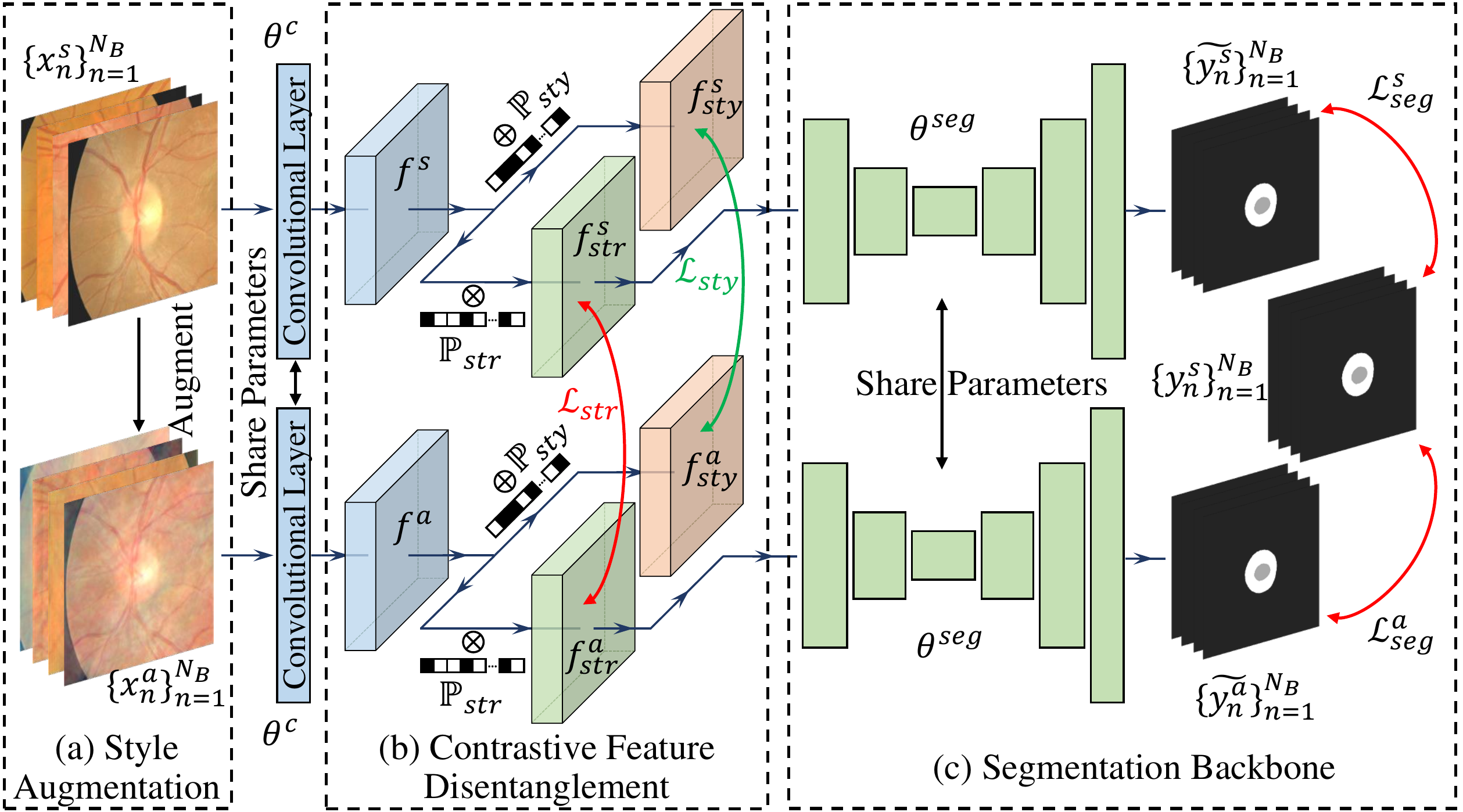}
  \caption{Diagram of our C$^2$SDG. The rectangles in blue and green represent the convolutional layer and the segmentation backbone, respectively. The cubes represent different features. The projectors with parameters $\theta^{p}$ in (b) are omitted for simplicity.}
  \label{fig:overview}
\end{figure}

\section{Method}
\subsection{Problem Definition and Method Overview}
Let the source domain be denoted by $\mathcal{D}^s=\{x_i^s, y_i^s\}_{i=1}^{N_s}$, 
where $x_i^s$ is the $i$-th source domain image,
and $y_i^s$ is its segmentation mask.
Our goal is to train a segmentation model $F_{\theta}:x\to y$ on $\mathcal{D}^s$, which can generalize well to an unseen target domain $\mathcal{D}^t=\{x_i^t\}_{i=1}^{N_t}$.
The proposed C$^2$SDG mainly consists of a segmentation backbone, a style augmentation (StyleAug) module, and a contrastive feature disentanglement (CFD) module.
For each image $x^s$, the StyleAug module generates its style-augmented counterpart $x^a$, which shares the same structure but different style to $x^s$.
Then a convolutional layer extracts high-dimensional representations $f^s$ and $f^a$ from $x^s$ and $x^a$.
After that, $f^s$ and $f^a$ are fed to the CFD module to perform contrastive training, resulting in the disentangled style representations $f_{sty}$ and structure representations $f_{str}$.
The segmentation backbone only takes $f_{str}$ as its input and generates the segmentation prediction $\widetilde{y}$.
Note that, although we take a U-shape network~\cite{falk2019u} as the backbone for this study, both StyleAug and CFD modules are modularly designed and can be incorporated into most segmentation backbones.
The diagram of our C$^2$SDG is shown in~\figurename{~\ref{fig:overview}}. 
We now delve into its details.

\subsection{Style Augmentation}
Given a batch of source domain data $\{x_n^s\}_{n=1}^{N_B}$,
we adopt a series of style-related data augmentation approaches, 
\ie, gamma correction and noise addition in BigAug~\cite{zhang2020generalizing}, and Bezier curve transformation in SLAug~\cite{su2022rethinking}, 
to generate $\{x_n^{BA}\}_{n=1}^{N_B}$ and $\{x_n^{SL}\}_{n=1}^{N_B}$.

Additionally, to fully utilize the style diversity inside single domain data, we also adopt low-frequency components replacement~\cite{yang_fda_2020} within a batch of source domain images.
Specifically, We reverse $\{x_n^s\}_{n=1}^{N_B}$ to match $x_n^s$ with $x_{r}^s$, where $r=N_B+1-n$ to ensure $x_{r}^s$ provides a different reference style.
Then we transform $x_n^s$ and $x_{r}^s$ to the frequency domain and exchange their low-frequency components $Low(Amp(x^s);\beta)$ in the amplitude map, where $\beta$ is the cut-off ratio between low and high-frequency components and is randomly selected from (0.05, 0.15].
After that, we recover all low-frequency exchanged images to generate $\{x_n^{FR}\}_{n=1}^{N_B}$.

The style-augmented images batch $\{x_n^{a}\}_{n=1}^{N_B}$ is set to $\{x_n^{BA}\}_{n=1}^{N_B}$, $\{x_n^{SL}\}_{n=1}^{N_B}$, and $\{x_n^{FR}\}_{n=1}^{N_B}$ in turn to perform contrastive training and segmentation.

\subsection{Contrastive Feature Disentanglement}
Given $x^s$ and $x^a$, 
we use a convolutional layer with parameter $\theta^c$ to generate their shallow features $f^s$ and $f^a$, which are 64-channel feature maps for this study.

Then we use a channel mask prompt $\mathbb{P}\in \mathbb{R}^{2\times64}$ to disentangle each shallow feature map $f$ into style representation $f_{sty}$ and structure representation $f_{str}$ explicitly channel-wisely
\begin{equation}
\begin{aligned}
\left\{\begin{matrix}
f_{sty}=f\times \mathbb{P}_{sty}=f\times SM(\frac{\mathbb{P}}{\tau})_1 \\ 
f_{str}=f\times \mathbb{P}_{str}=f\times SM(\frac{\mathbb{P}}{\tau})_2,
\end{matrix}\right.
\end{aligned}
\end{equation}
where $SM(\cdot)$ is a softmax function, the subscript i denotes i-th channel, and $\tau=0.1$ is a temperature factor that encourages $\mathbb{P}_{sty}$ and $\mathbb{P}_{str}$ to be binary-element vectors, \ie, approximately belonging to $\{0,1\}^{64}$.

After channel-wise feature disentanglement, we have $\{f_{sty}^s, f_{str}^s\}$ from $x^s$ and $\{f_{sty}^a, f_{str}^a\}$ from $x^a$.
It is expected that 
(a) $f_{sty}^s$ and $f_{sty}^a$ are different since we want to identify them as the style-sensitive `devil' channels,
and (b) $f_{str}^s$ and $f_{str}^a$ are the same since we want to identify them as the style-irrelevant channels and $x^s$ and $x^a$ share the same structure.
Therefore, we design two contrastive loss functions \textcolor{mygreen}{$\mathcal{L}_{sty}$} and \textcolor{myred}{$\mathcal{L}_{str}$}
\begin{equation}
\begin{aligned}
\left\{\begin{matrix}
\textcolor{myred}{\mathcal{L}_{str}} = \sum|Proj(f_{str}^s) - Proj(f_{str}^a)| \\ 
\textcolor{mygreen}{\mathcal{L}_{sty}} = -\sum|Proj(f_{sty}^s) - Proj(f_{sty}^a)|,
\end{matrix}\right.
\end{aligned}
\end{equation}
where the $Proj(\cdot)$ with parameters $\theta^{p}$ reduces the dimension of $f_{str}$ and $f_{sty}$.

Only $f_{str}^s$ and $f_{str}^a$ are fed to the segmentation backbone with parameters $\theta^{seg}$ to generate the segmentation predictions $\widetilde{y^s}$ and $\widetilde{y^a}$.

\subsection{Training and Inference}
\noindent \textbf{Training.}
For the segmentation task, we treat OC/OD segmentation as two binary segmentation tasks and adopt the binary cross-entropy loss as our objective
\begin{equation}
\begin{aligned}
\mathcal{L}_{ce}(y, \widetilde{y})
= -(\widetilde{y} \log y +\left(1-\widetilde{y}\right) \log \left(1-y\right))
\end{aligned}
\end{equation}
where $y$ represents the segmentation ground truth and $\widetilde{y}$ is the prediction.
The total segmentation loss can be calculated as
\begin{equation}
\begin{aligned}
\mathcal{L}_{seg}= \mathcal{L}_{seg}^s + \mathcal{L}_{seg}^a
= \mathcal{L}_{ce}(y^s, \widetilde{y^s}) + \mathcal{L}_{ce}(y^s, \widetilde{y^a}).
\end{aligned}
\end{equation}

During training, we alternately minimize $\mathcal{L}_{seg}$ to optimize $\{\theta^c,\mathbb{P},\theta^{seg}\}$, and minimize $\textcolor{myred}{\mathcal{L}_{str}} + \textcolor{mygreen}{\mathcal{L}_{sty}}$ to optimize $\{\mathbb{P},\theta^{p}\}$.

\noindent \textbf{Inference.}
Given a test image $x^t$, its shallow feature map $f^t$ can be extracted by the first convolutional layer. 
Based on $f^t$, the optimized channel mask prompt $\mathbb{P}$ can separate it into $f_{sty}^t$ and $f_{str}^t$. 
Only $f_{str}^t$ is fed to the segmentation backbone to generate the segmentation prediction $\widetilde{y^t}$.

\section{Experiments and Results}
\noindent \textbf{Materials and Evaluation Metrics.}
The multi-domain joint OC/OD segmentation dataset RIGA+~\cite{hu_domain_2022,decenciere2014feedback,almazroa2018retinal} was used for this study.
It contains annotated fundus images from five domains, including 
195 images from BinRushed, 
95 images from Magrabia, 
173 images from BASE1,
148 images from BASE2, and 
133 images from BASE3.
Each image was annotated by six raters, and only the first rater's annotations were used in our experiments.
We chose BinRushed and Magrabia, respectively, as the source domain to train the segmentation model, and evaluated the model on the other three (target) domains.
We adopted the Dice Similarity Coefficient ($D$, \%) to measure the segmentation performance. 

\noindent \textbf{Implementation Details.}
The images were center-cropped and normalized by subtracting the mean and dividing by the standard deviation.
The input batch contains eight images of size $512\times512$.
The U-shape segmentation network, whose encoder is a modified ResNet-34, was adopted as the segmentation backbone of our C$^2$SDG and all competing methods for a fair comparison.
The projector in our CFD module contains a convolutional layer followed by a batch normalization layer, a max pooling layer, and a fully connected layer to convert $f_{sty}$ and $f_{str}$ to 1024-dimensional vectors.
The SGD algorithm with a momentum of 0.99 was adopted as the optimizer.
The initial learning rate was set to $lr_0=0.01$ and decayed according to $lr=lr_0\times(1-e/E)^{0.9}$, where $e$ is the current epoch and $E=100$ is the maximum epoch.
All experiments were implemented using the PyTorch framework and performed with one NVIDIA 2080Ti GPU.

\begin{figure}[t]
  \centering
  \includegraphics[width=1\textwidth]{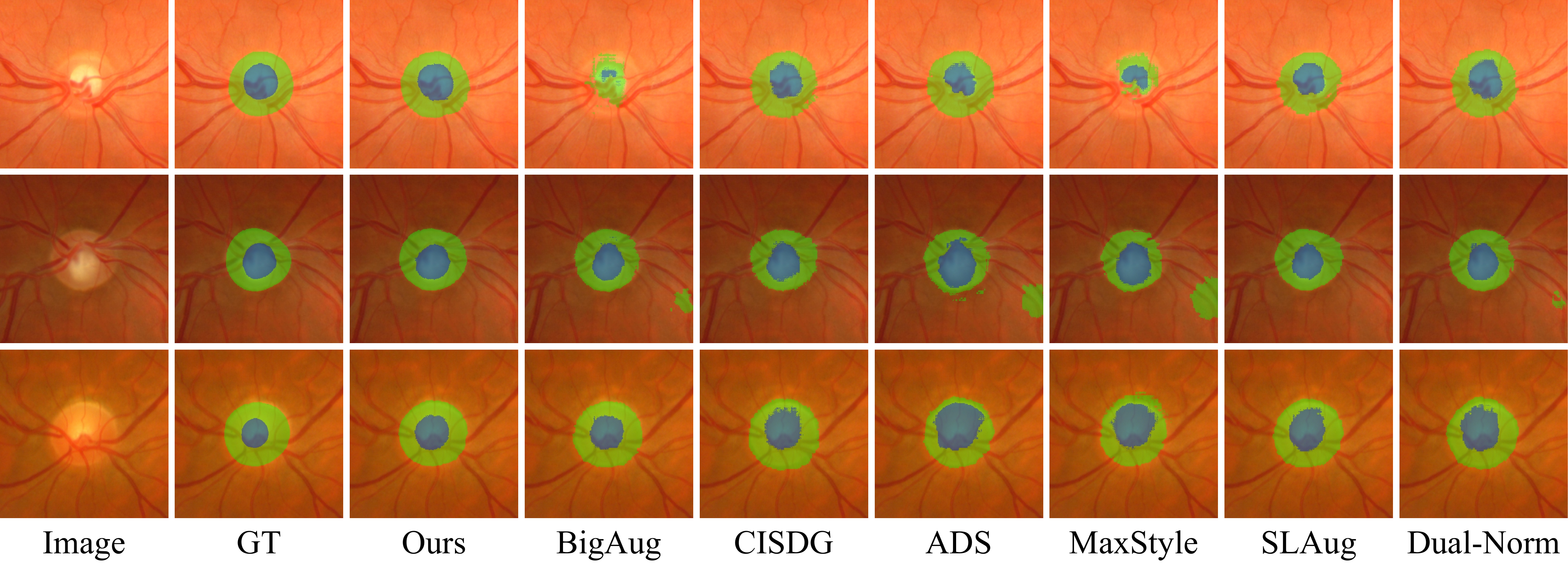}
  \caption{Visualization of segmentation masks predicted by our C$^2$SDG and six competing methods, together with ground truth.}
  \label{fig:compare}
\end{figure}

\noindent \textbf{Comparative Experiments.}
We compared our C$^2$SDG with two baselines, 
including `Intra-Domain' (\ie, training and testing on the data from the same target domain using 3-fold cross-validation) 
and `w/o SDG' (\ie, training on the source domain and testing on the target domain),
and six SDG methods, including 
BigAug~\cite{zhang2020generalizing}, CISDG~\cite{ouyang2022causality}, ADS~\cite{xu2022adversarial}, MaxStyle~\cite{chen2022maxstyle}, SLAug~\cite{su2022rethinking}, and Dual-Norm~\cite{zhou2022generalizable}.
In each experiment, only one source domain is used for training, ensuring that only the data from a single source domain can be accessed during training.
For a fair comparison, all competing methods are re-implemented using the same backbone as our C$^2$SDG based on their published code and paper.
The results of C$^2$SDG and its competitors were given in~\tablename{~\ref{tab:compare}}.
It shows that C$^2$SDG improves the performance of `w/o SDG' with a large margin and outperforms all competing SDG methods.
We also visualize the segmentation predictions generated by our C$^2$SDG and six competing methods in~\figurename{~\ref{fig:compare}}. 
It reveals that our C$^2$SDG can produce the most accurate segmentation map.

\begin{table}[t]
\renewcommand\arraystretch{0.9}
\centering
\caption{
Average performance of three trials of our C$^2$SDG and six competing methods in joint OC/OD segmentation using BinRushed (row 2 $\sim$ row 9) and Magrabia (row 10 $\sim$ row 17) as source domain, respectively. Their standard deviations are reported as subscripts. The performance of `Intra-Domain' and `w/o SDG' is displayed for reference.
The best results except for `Intra-Domain' are highlighted in \textcolor{blue}{blue}.
}
\label{tab:compare}
\begin{tabular}{l|c|c|c|c|c|c|c|c}
\hline
\hline
\multicolumn{1}{c|}{\multirow{2}{*}{Methods}}                         & \multicolumn{2}{c|}{BASE1}                              & \multicolumn{2}{c|}{BASE2}                              & \multicolumn{2}{c|}{BASE3}                           & \multicolumn{2}{c}{Average}                                \\ \cline{2-9} 
\multicolumn{1}{c|}{}                         & \multicolumn{1}{c|}{$D_{OD}$}    & \multicolumn{1}{c|}{$D_{OC}$}    & \multicolumn{1}{c|}{$D_{OD}$}    & \multicolumn{1}{c|}{$D_{OC}$}    & \multicolumn{1}{c|}{$D_{OD}$}    & \multicolumn{1}{c|}{$D_{OC}$} & \multicolumn{1}{c|}{$D_{OD}$}    & \multicolumn{1}{c}{$D_{OC}$} \\ \hline
\hline
Intra-Domain                           & $94.71_{0.07}$ & $84.07_{0.35}$ & $94.84_{0.18}$ & $86.32_{0.14}$ & $95.40_{0.05}$ & $87.34_{0.11}$ & 94.98 & 85.91 \\ \hline
\hline
w/o SDG                                & $91.82_{0.54}$ & $77.71_{0.88}$ & $79.78_{2.10}$ & $65.18_{3.24}$ & $88.83_{2.15}$ & $75.29_{3.23}$ & 86.81 & 72.73 \\ \hline
BigAug~\cite{zhou2022generalizable}    & $94.01_{0.34}$ & $81.51_{0.58}$ & $85.81_{0.68}$ & $71.12_{1.64}$ & $92.19_{0.51}$ & $79.75_{1.44}$ & 90.67 & 77.46 \\ \hline
CISDG~\cite{ouyang2022causality}       & $93.56_{0.13}$ & $81.00_{1.01}$ & $94.38_{0.23}$ & $83.79_{0.58}$ & $93.87_{0.03}$ & $83.75_{0.89}$ & 93.93 & 82.85 \\ \hline
ADS~\cite{xu2022adversarial}           & $94.07_{0.29}$ & $79.60_{5.06}$ & $94.29_{0.38}$ & $81.17_{3.72}$ & $93.64_{0.28}$ & $81.08_{4.97}$ & 94.00 & 80.62 \\ \hline
MaxStyle~\cite{chen2022maxstyle}       & $94.28_{0.14}$ & $82.61_{0.67}$ & $86.65_{0.76}$ & $74.71_{2.07}$ & $92.36_{0.39}$ & $82.33_{1.24}$ & 91.09 & 79.88 \\ \hline
SLAug~\cite{su2022rethinking}          & $95.28_{0.12}$ & $83.31_{1.10}$ & $95.49_{0.16}$ & $81.36_{2.51}$ & \textcolor{blue}{$95.57_{0.06}$} & $84.38_{1.39}$ & 95.45 & 83.02 \\ \hline
Dual-Norm~\cite{zhou2022generalizable} & $94.57_{0.10}$ & $81.81_{0.76}$ & $93.67_{0.11}$ & $79.16_{1.80}$ & $94.82_{0.28}$ & $83.67_{0.60}$ & 94.35 & 81.55 \\ \hline
Ours                                   & \textcolor{blue}{$95.73_{0.08}$} & \textcolor{blue}{$86.13_{0.07}$} & \textcolor{blue}{$95.73_{0.09}$} & \textcolor{blue}{$86.82_{0.58}$} & $95.45_{0.04}$ & \textcolor{blue}{$86.77_{0.19}$} & \textcolor{blue}{95.64} & \textcolor{blue}{86.57} \\ \hline
\hline
w/o SDG                                & $89.98_{0.54}$ & $77.21_{1.15}$ & $85.32_{1.79}$ & $73.51_{0.67}$ & $90.03_{0.27}$ & $80.71_{0.63}$ & 88.44 & 77.15 \\ \hline
BigAug~\cite{zhou2022generalizable}    & $92.32_{0.13}$ & $79.68_{0.38}$ & $88.24_{0.82}$ & $76.69_{0.37}$ & $91.35_{0.14}$ & $81.43_{0.78}$ & 90.64 & 79.27 \\ \hline
CISDG~\cite{ouyang2022causality}       & $89.67_{0.76}$ & $75.39_{3.22}$ & $87.97_{1.04}$ & $76.44_{3.48}$ & $89.91_{0.64}$ & $81.35_{2.81}$ & 89.18 & 77.73 \\ \hline
ADS~\cite{xu2022adversarial}           & $90.75_{2.42}$ & $77.78_{4.23}$ & $90.37_{2.07}$ & $79.60_{3.34}$ & $90.34_{2.93}$ & $79.99_{4.02}$ & 90.48 & 79.12 \\ \hline
MaxStyle~\cite{chen2022maxstyle}       & $91.63_{0.12}$ & $78.74_{1.95}$ & $90.61_{0.45}$ & $80.12_{0.90}$ & $91.22_{0.07}$ & $81.90_{1.14}$ & 91.15 & 80.25 \\ \hline
SLAug~\cite{su2022rethinking}          & $93.08_{0.17}$ & $80.70_{0.35}$ & $92.70_{0.12}$ & $80.15_{0.43}$ & $92.23_{0.16}$ & $80.89_{0.14}$ & 92.67 & 80.58 \\ \hline
Dual-Norm~\cite{zhou2022generalizable} & $92.35_{0.37}$ & $79.02_{0.39}$ & $91.23_{0.29}$ & $80.06_{0.26}$ & $92.09_{0.28}$ & $79.87_{0.25}$ & 91.89 & 79.65 \\ \hline
Ours & \textcolor{blue}{$94.78_{0.03}$} & \textcolor{blue}{$84.94_{0.36}$} & \textcolor{blue}{$95.16_{0.09}$} & \textcolor{blue}{$85.68_{0.28}$} & \textcolor{blue}{$95.00_{0.09}$} & \textcolor{blue}{$85.98_{0.29}$} & \textcolor{blue}{94.98} & \textcolor{blue}{85.53} \\ \hline
\hline
\end{tabular}
\end{table}

\noindent \textbf{Ablation Analysis.}
To evaluate the effectiveness of low-frequency components replacement (FR) in StyleAug and CFD, we conducted ablation experiments using BinRushed and Magrabia as the source domain, respectively. 
The average performance is shown in~\tablename{~\ref{tab:ablation}}.
The performance of using both BigAug and SLAug is displayed as `Baseline'.
It reveals that both FR and CFD contribute to performance gains.

\noindent \textbf{Analysis of CFD.}
Our CFD is modularly designed and can be incorporated into other SDG methods.
We inserted our CFD to ADS~\cite{xu2022adversarial} and SLAug~\cite{su2022rethinking}, respectively.
The performance of these two approaches and their variants, denoted as C$^2$-ADS and C$^2$-SLAug, was shown in~\tablename{~\ref{tab:modular}}.
It reveals that our CFD module can boost their ability to disentangle structure representations and improve the segmentation performance on the target domain effectively.
We also adopted `Ours w/o CFD' as `Baseline' and compared the channel-level contrastive feature disentanglement strategy with the adversarial training strategy and channel-level dropout (see~\tablename{~\ref{tab:contrastive}}).
It shows that the adversarial training strategy fails to perform channel-level feature disentanglement,
due to the limited training data~\cite{clarysse2023why} for SDG.
Nonetheless, our channel-level contrastive learning strategy achieves the best performance compared to other strategies, further confirming the effectiveness of our CFD module.

\noindent
\begin{minipage}{\textwidth}
\begin{minipage}[t]{0.32\textwidth}
\makeatletter\def\@captype{table}
\setlength{\belowcaptionskip}{7pt}
\renewcommand\arraystretch{0.9}
\centering
\setlength\tabcolsep{3pt}
\caption{
Average performance of our C$^2$SDG and three variants.
}
\label{tab:ablation}
\begin{tabular}{l|c|c}
\hline
\hline
\multicolumn{1}{c|}{\multirow{2}{*}{Methods}} & \multicolumn{2}{c}{Average}                        \\ \cline{2-3} 
\multicolumn{1}{c|}{}                         & \multicolumn{1}{c|}{$D_{OD}$}   & \multicolumn{1}{c}{$D_{OC}$} \\ \hline
\hline
Baseline & 94.13 & 81.62 \\ \hline
w/o FR & 95.07 & 84.90 \\ \hline
w/o CFD & 95.07 & 84.83 \\ \hline
Ours & 95.31 & 86.05 \\ \hline
\hline
\end{tabular}
\end{minipage}
\begin{minipage}[t]{0.34\textwidth}
\makeatletter\def\@captype{table}
\setlength{\belowcaptionskip}{7pt}
\renewcommand\arraystretch{0.9}
\centering
\setlength\tabcolsep{3pt}
\caption{
Average performance of ADS, SLAug, and their two variants.
}
\label{tab:modular}
\begin{tabular}{l|c|c}
\hline
\hline
\multicolumn{1}{c|}{\multirow{2}{*}{Methods}} & \multicolumn{2}{c}{Average}                        \\ \cline{2-3} 
\multicolumn{1}{c|}{}                         & \multicolumn{1}{c|}{$D_{OD}$}   & \multicolumn{1}{c}{$D_{OC}$} \\ \hline
\hline
ADS~\cite{xu2022adversarial} & 92.24 & 79.87 \\ \hline
C$^2$-ADS & 93.76 & 81.35 \\ \hline
SLAug~\cite{su2022rethinking} & 94.06 & 81.80 \\ \hline
C$^2$-SLAug & 94.24 & 83.68 \\ \hline
\hline
\end{tabular}
\end{minipage}
\begin{minipage}[t]{0.32\textwidth}
\makeatletter\def\@captype{table}
\setlength{\belowcaptionskip}{7pt}
\renewcommand\arraystretch{0.9}
\centering
\setlength\tabcolsep{3pt}
\caption{
Average performance of using contrastive and other strategies.
}
\label{tab:contrastive}
\begin{tabular}{l|c|c}
\hline
\hline
\multicolumn{1}{c|}{\multirow{2}{*}{Methods}} & \multicolumn{2}{c}{Average}                        \\ \cline{2-3} 
\multicolumn{1}{c|}{}                         & \multicolumn{1}{c|}{$D_{OD}$}   & \multicolumn{1}{c}{$D_{OC}$} \\ \hline
\hline
Baseline & 95.07 & 84.83 \\ \hline
Dropout & 95.14 & 84.95 \\ \hline
Adversarial & 90.27 & 78.47 \\ \hline
Ours & 95.31 & 86.05 \\ \hline
\hline
\end{tabular}
\end{minipage}
\end{minipage}

\section{Conclusion}
In this paper, we propose a novel SDG method called C$^2$SDG for medical image segmentation.
In C$^2$SDG, the StyleAug module generates style-augmented counterpart of each source domain image and enables contrastive learning, the CFD module performs channel-level style and structure representations disentanglement via optimizing a channel prompt $\mathbb{P}$, and the segmentation is performed based solely on structure representations.
Our results on a multi-domain joint OC/OD segmentation benchmark indicate the effectiveness of StyleAug and CFD and also suggest that our C$^2$SDG outperforms the baselines and six completing SDG methods with a large margin.

 \subsubsection{Acknowledgement:} 
This work was supported in part by the National Natural Science Foundation of China under Grant 62171377, in part by the Key Technologies Research and Development Program under Grant 2022YFC2009903 / 2022YFC2009900, in part by the Natural Science Foundation of Ningbo City, China, under Grant 2021J052, and in part by the Innovation Foundation for Doctor Dissertation of Northwestern Polytechnical University under Grant CX2023016.

% \clearpage
\bibliographystyle{splncs04}
\bibliography{reference}

\clearpage

\title{Supplementary for Contrastive Single Domain Generalization for Medical Image Segmentation}
\titlerunning{Contrastive Single Domain Generalization for Medical Image Segmentation}
\author{Shishuai Hu\inst{1} \and
Zehui Liao\inst{1} \and
Yong Xia\inst{1,2,3}\Envelope}
\authorrunning{S. Hu et al.}
\institute{
National Engineering Laboratory for Integrated Aero-Space-Ground-Ocean Big Data Application Technology, School of Computer Science and Engineering, Northwestern Polytechnical University, Xi’an 710072, China \\
\email{yxia@nwpu.edu.cn}
\and
Ningbo Institute of Northwestern Polytechnical University, Ningbo 315048, China
\and
Research and Development Institute of Northwestern Polytechnical University in Shenzhen, Shenzhen 518057, China
}
\maketitle              % typeset the header of the contribution

\begin{table}[]
\setlength\tabcolsep{5pt}
% \renewcommand\arraystretch{1.05}
% \scriptsize
\centering
\caption{
Average performance of the combination of different data augmentation methods. 
`SDAug' represents structure-destroyed augmentation, such as rotation, elastic transform, and scaling. 
`BigAug' contains Gaussian noise addition, Gamma transform, and Gaussian blur transform.
`SLAug' represents Bezier curve transform.
`FRAug' denotes low-frequency component replacement.
}
\label{tab:styleaug-ablation}
\begin{tabular}{c|c|c|c|c|c|c|c|c|c}
\hline
\hline
\multicolumn{4}{c|}{Augmentation Methods}  & \multicolumn{2}{c|}{BinRusheda} & \multicolumn{2}{c|}{Magrabia} & \multicolumn{2}{c}{Average} \\ \hline
SDAug & BigAug & SLAug & FRAug & $D_{OD}$ & $D_{OC}$ & $D_{OD}$ & $D_{OC}$ & $D_{OD}$ & $D_{OC}$  \\ \hline
\hline
           &            &            &            & 83.45 & 69.17 & 88.76 & 77.39 & 86.11 & 73.28 \\ \hline
\checkmark &            &            &            & 86.81 & 72.73 & 88.44 & 77.15 & 87.63 & 74.94 \\ \hline
\checkmark & \checkmark &            &            & 90.67 & 77.46 & 90.64 & 79.27 & 90.66 & 78.37 \\ \hline
\checkmark & \checkmark & \checkmark &            & 94.90 & 82.67 & 93.36 & 80.56 & 94.13 & 81.62 \\ \hline
\checkmark & \checkmark &            & \checkmark & 94.17 & 83.28 & 93.71 & 82.20 & 93.94 & 82.74 \\ \hline
\checkmark & \checkmark & \checkmark & \checkmark & 95.73 & 85.39 & 94.40 & 84.27 & 95.07 & 84.83 \\ \hline
\hline
\end{tabular}
\end{table}

\begin{figure}[]
  \centering
  \includegraphics[width=1\textwidth]{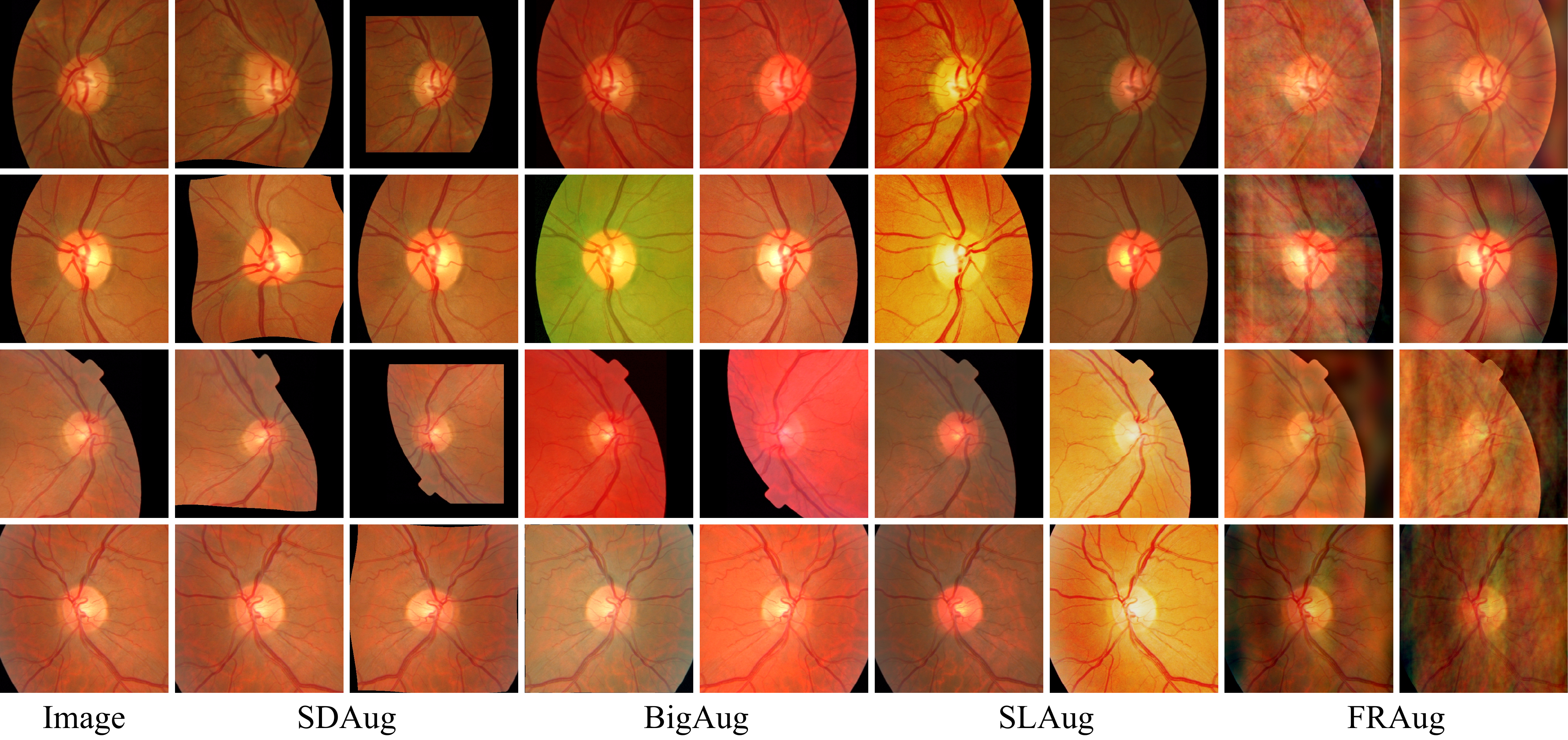}
  \caption{Illustration of the augmented images of different augmentation methods.}
  \label{fig:augmentation}
\end{figure}

\begin{table}[]
\setlength\tabcolsep{4pt}
% \renewcommand\arraystretch{1.05}
% \scriptsize
\centering
\caption{
Performance of our C$^2$SDG when using different channel prompt initialization methods (the first block) and using different projectors (the second block).
}
\label{tab:cfd-ablation}
\begin{tabular}{c|c|c|c|c|c|c|c}
\hline
\hline
\multicolumn{2}{c|}{\multirow{2}{*}{Methods}}                                    & \multicolumn{2}{c|}{BinRusheda}                      & \multicolumn{2}{c|}{Magrabia}                        & \multicolumn{2}{c}{Average}                         \\ \cline{3-8} 
\multicolumn{2}{c|}{}                                                            & $D_{OD}$ & $D_{OC}$ & $D_{OD}$ & $D_{OC}$ & $D_{OD}$ & $D_{OC}$ \\ \hline
\hline
\multicolumn{1}{c|}{\multirow{2}{*}{Prompt Initialization}} & 1-0 Initialization & 95.92 & 85.79                   & 95.01 & 85.09                   & 95.47 & 85.44                  \\ \cline{2-8} 
                                                            & Random             & 95.64 & 86.57                   & 94.98 & 85.53                   & 95.31 & 86.05                  \\ \hline
\hline
\multicolumn{1}{c|}{\multirow{2}{*}{Projector}}             & No Projector       & 95.15 & 84.58                   & 94.94 & 85.42                   & 95.05 & 85.00                  \\ \cline{2-8} 
                                                            & Our design         & 95.64 & 86.57                   & 94.98 & 85.53                   & 95.31 & 86.05                  \\ \hline
\hline
\end{tabular}
\end{table}

\begin{figure}[]
  \centering
  \includegraphics[width=1\textwidth]{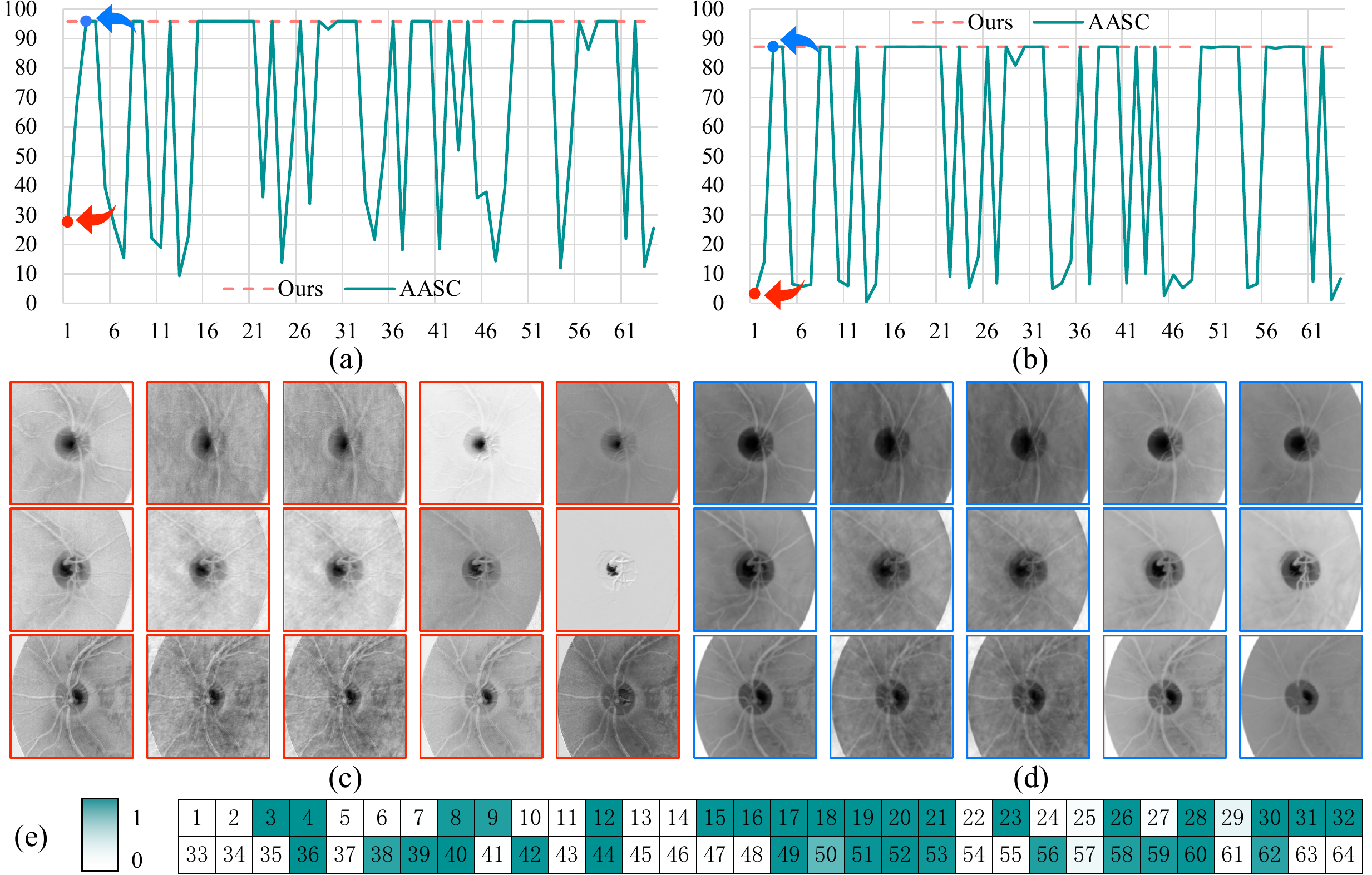}
  \caption{The average segmentation performance of the model for (a) OD and (b) OC on the unseen target domain (BASE2). 
The impact of adding a specific channel in $f_{sty}$ on C$^2$SDG's segmentation performance is illustrated by `AASC'.
The performance of our C$^2$SDG is displayed as `Ours'.
Feature visualizations for the 1st and 3rd channels when fed three target images and their style-augmented counterparts are shown in (c) and (d).
The optimized $\mathbb{P}_{str}$ is illustrated in (e).}
  \label{fig:results}
\end{figure}

\end{document}